\documentclass[runningheads]{llncs}

\usepackage{eccv}

\usepackage{eccvabbrv}

\usepackage{graphicx}
\usepackage{booktabs}
\usepackage{wrapfig}

\usepackage[accsupp]{axessibility}  

\usepackage[pagebackref,breaklinks,colorlinks,citecolor=eccvblue]{hyperref}

\usepackage{orcidlink}

\begin{document}

\title{MUL-T: Decoding Spatial Cellular Architecture in Multiplexed Tissue Images} 

\titlerunning{MUL-T}

\author{Farzaneh Seyedshahi\inst{1,2}\orcidlink{0009-0007-1111-052X} \and
Kai Rakovic\inst{1,2}\orcidlink{0000-0002-0676-1973} \and
Adalberto Claudio Quiros \inst{2}\orcidlink{0000-0003-4804-0741} \and
John LeQuesne\inst{1,2}\orcidlink{0000-0003-3552-7446}\and
Ke Yuan\inst{1,2}\orcidlink{0000-0002-2318-1460}
}

\authorrunning{F.Seyedshahi et al.}

\institute{Cancer Research UK Scotland Institute, Glasgow, UK \and
University of Glasgow, Glasgow, UK
}

\maketitle

\begin{abstract}
Understanding tissue organisation in multiplexed imaging requires modelling both cellular phenotypes and their spatial context. Existing approaches typically rely on handcrafted features, such as marker intensity statistics or cell-type proportions, which often fail to scale or generalise across cohorts with heterogeneous marker panels. We introduce MUL-T, a lightweight transformer framework that reframes tissue architecture as a masked contextual prediction task over discrete cell tokens. By learning contextualised [CLS] embeddings without task-specific supervision, the model captures higher-order cellular interactions while remaining computationally efficient. We evaluate MUL-T on several clinically relevant downstream tasks, including core-level tumour pattern classification, patient-level grading, PD-L1 positivity prediction, and cross-dataset treatment response prediction. Across tasks, MUL-T consistently outperforms classical feature-based baselines and achieves performance comparable to a foundation ViT model, despite substantially fewer parameters and lower training cost.
\end{abstract}

\section{Introduction}
\label{sec:intro}
Spatially resolved tissue imaging has become a key modality for studying complex biological systems, enabling the measurement of cellular phenotypes along with their spatial organisation at single-cell resolution \cite{de2024multiplex, janeiro2024spatially}. Beyond individual cell states, the arrangement and co-occurrence of cells encode critical information about tissue structure and function \cite{hanahan2022hallmarks, hanahan2011hallmarks}. From a computational perspective, this creates a structured data problem where meaning arises not from isolated observations, but from spatial context and interactions across many heterogeneous entities. Effectively modelling such spatial context remains a central challenge\cite{ fassler2020deep}.
Recent advances in multiplexed imaging technologies have substantially increased both the dimensionality and scale of spatial tissue profiling \cite{de2024multiplex, lin2015highly}. Modern platforms routinely generate datasets comprising millions of cells, each described by high-dimensional protein marker vectors and precise spatial coordinates \cite{wang2023spatial}. In cancer, such data provides an unprecedented opportunity to interrogate tumour microenvironment organisation, cellular neighbourhood structure, and rare or spatially restricted phenotypes that are not apparent from bulk measurements alone. However, heterogeneous cell states, dense spatial packing, and multi-channel signals create a level of combinatorial complexity that current computational pipelines are not designed to handle. 

Traditional analysis of multiplexed tissue imaging typically relies on marker intensity and manual thresholding, by platforms like QuPath\cite{bankhead2017qupath}. These methods use rigid gating strategies and effectively treat cells as binary (positive or negative) for specific proteins. While computationally simple, this approach is highly sensitive to signal noise and batch effects. Also, because it requires subjective manual tuning, it limits both automation and reproducibility across different studies.\cite{fassler2020deep, garbo2025evaluating}. 

Another common paradigm uses algorithms like PhenoGraph\cite{levine2015data} for high-dimensional clustering and aggregation of the clustered cells at tissue level. While effective at identifying phenotypic states, these tools often adopt a "bag-of-cells" mentality. By focusing on global cluster frequencies, they decontextualise the data and strip away the spatial coordinates that define tissue structure.

Modern graph-based formulations often assume local homophily, meaning neighbouring nodes (cells) are similar\cite{wu2022graph}. However, tissue environments are inherently heterogeneous; critical interactions often occur between dissimilar or rare cell types. These signals can be diluted by standard neighbourhood aggregation, which tends to be dominated by the most abundant populations\cite{yun2024mew}. Furthermore, constructing and processing cell-level graphs for large tissue samples is computationally expensive, often requiring aggressive subsampling or neighbourhood truncation, which limits the ability to capture long-range spatial structure and undermines scalability to large cohorts.

Vision Transformer (ViT) architectures have emerged as a robust alternative to traditional methods, using attention mechanisms to model both local structures and long-range dependencies. These models have demonstrated strong representation learning in natural imagery \cite{oquab2023dinov2, radford2021learning} and histopathology\cite{chen2024uni, xu2024whole, vorontsov2024foundation}. Typically, the pipeline partitions whole-slide images into fixed size pixel tiles (\eg \(256\times256\)), encode them independently, and aggregate embeddings using strategies like Multiple Instance Learning (MIL) or different pooling methods. While effective for RGB histopathology images, this paradigm is less suited for multiplexed imaging for several reasons. The high-dimensional marker channels in mIF are difficult to interpret simultaneously. Collapsing channels into an RGB-pixel tile entangles spatial structure with marker semantics, making it hard to distinguish phenotypic variation from spatial context. Moreover, ViTs operate on fixed grids and assume consistent channel semantics. However, mIF marker panels vary significantly across cohorts and platforms, preventing the direct transfer of pretrained models on a cohort to another one. Also, training panel-agnostic foundation models requires massive, harmonised datasets that are currently scarce and expensive to produce. Finally and crucially, tile-based embeddings often blur cell boundaries and dilute signals from rare cell populations or specific microniches; both of which are critical to cancer biology and discovery.
 
These limitations motivate a modelling framework that separates marker-space reasoning from spatial reasoning and shifting the focus from patch-level to cell-level analysis. By decoupling these elements, the model can more effectively support transfer learning across heterogeneous panels, allowing knowledge gained from large primary cohorts to generalise to smaller, specialised datasets without sacrificing biologically meaningful resolution. Furthermore, by modelling at the cellular level rather than processing dense pixel grids, we can significantly reduce the overhead required for large-scale deployments. Focusing on cell-level modelling also addresses the high cost of data acquisition, aiming for a robust architecture that maintains high performance even when training on the smaller, more limited datasets typically available in clinical settings.

In this work, we formulate spatial cellular organisation as a contextual modelling problem by representing tissue as sequences of discrete tokens. Drawing inspiration from natural language, each segmented cell constitutes a token, with spatially contiguous regions defining sequences similar to sentences. This shift from pixel-based to biologically grounded entities uses cell segmentation and unsupervised phenotypic clustering to create a compact, interpretable abstraction of high-dimensional marker profiles. The cell tokens are granular enough to capture rare phenotypes while maintaining global structural logic.

To process these sequences, we employ a lightweight transformer that operates directly on cell-level tokens. Through self-attention, the model learns spatial dependencies and produces a contextualised \texttt{[CLS]} embedding that summarises tissue architecture for downstream tasks. By opting for a tiny transformer architecture, the framework achieves high performance without the need for massive, foundation-scale datasets. Furthermore, modelling at the cellular level provides a higher density of training tokens compared to patch-based approaches, improving statistical efficiency and preserving biological resolution. This balance ensures scalable training and robust cross-dataset transfer, resulting a reusable model that generalises across heterogeneous cohorts and downstream tasks.

\section{Methodology}
\begin{wrapfigure}[14]{r}{.35\textwidth}
  \centering
  \includegraphics[width=.35\textwidth]{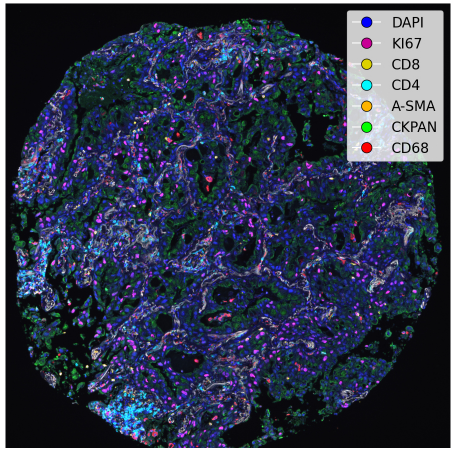}
  \caption{Example mIF image from the LATTICeA-IO cohort.}
  \label{fig:mif}
\end{wrapfigure}

In this section, we describe the two datasets used in this study and formalise the problem setting for tokenisation, transformer training, and downstream evaluation. \Cref{fig:mif} shows a representative multiplex tissue microarray (TMA) core from the primary dataset, where individual marker channels are colour-coded and composited into an RGB visualisation.

\subsection{Datasets}

\subsubsection{LATTICeA-IO (Primary Dataset)}

The primary dataset for this study is LATTICeA-IO, a large-scale multiplex immunofluorescence (mIF) cohort. It comprises approximately 2,500 TMA cores (1mm diameter) from 1,025 patients with lung adenocarcinoma, representing a total of roughly 17 million segmented cells. 

The marker panel is designed to capture key axes of the tumour microenvironment using spectrally distinct Opal fluorophores. These include: DAPI (nuclear counterstain), Ki-67 (proliferation), CD68 (macrophages), CD8 and CD4 (cytotoxic and helper T cells), $\alpha$-SMA (stromal/myofibroblastic components), and pan-cytokeratin (epithelial compartment). Whole-slide images were acquired at 20x magnification and spectrally unmixed via Phenochart to generate quantitative per-channel fluorescence maps. For downstream tasks, the dataset provides core-level labels (including predominant tumour patterns and PD-L1 positivity) alongside patient-level clinical gradings. Cell segmentation was performed by identifying nuclei in the DAPI channel followed by whole-cell expansion to approximate cytoplasmic boundaries, enabling extraction of per-cell marker intensities across nuclear and non-nuclear compartments. The resulting dataset provides high-dimensional, spatially resolved single-cell representations spanning epithelial, immune, proliferative, and stromal niches, forming the foundation for downstream tokenisation and transformer-based modelling.


\subsubsection{CTCL-P (Additional Dataset)}
External validation was performed using a public mIF dataset\cite{phillips2021immune}, which profiles cutaneous T-cell lymphoma (CTCL) samples treated with pembrolizumab using CODEX\cite{black2021codex} imaging. This dataset contains 70 tumour cores from 14 patients and includes a panel of approximately 56 protein markers. Several markers overlap biologically with the LATTICeA-IO panel, including CD8, CD4, Ki-67, and CD68. For markers not directly available, biologically analogous substitutes were used: DRAQ5 served as a nuclear marker in place of DAPI, and cytokeratin markers were used to represent epithelial components in place of CKPAN. To enable cross-dataset generalisation, a shared-marker alignment protocol was used to transfer token identities learned from the primary dataset to the CTCL-P dataset (see \cref{crossdatasetassignment}). The downstream task on CTCL-P is binary classification of responders versus non-responders to PD-1 blockade based solely on generated \texttt{[CLS]} tokens from the transformer trained on primary dataset. In total, the dataset contains approximately 117,000 cells.

\subsection{Problem formulation}
\label{problemformulation}
The objective of this work is to model spatial cellular organisation in histological tissue by drawing an analogy to natural language. In natural language, semantic meaning emerges from contextual relationships between words within a sentence. Similarly, we aim to learn representations of tissue architecture from the spatial context and phenotypic identity of cells, thereby defining a language of cells. In this language, individual cells are treated as discrete tokens, and spatially localised groups of cells form fixed-size windows across the tissue.

The proposed framework consists of two main stages. First, in the tokenisation step, cells are tokenised by assigning discrete phenotypic and spatial identifiers. This stage also includes cross-dataset cluster assignment by supporting consistent tokenisation to the additional dataset with different marker panels. Second, sequences of cell tokens are used to train a transformer-based language model that captures spatial dependencies through masked token prediction. Once trained, the model produces contextualised sequence-level representations via a classification (CLS) token, which summarises spatial relationships within each window or sequence. These \texttt{[CLS]} embeddings are subsequently pooled as representations of tissue cores or patients for downstream prediction tasks.

\subsubsection{Tokenisation}
To tokenise cells, mIF images are processed using a U-Net-based \cite{ronneberger2015u} cell segmentation framework. Nuclei are identified via the DAPI marker, a standard approach for reliable nucleus detection. For every detected cell, precise spatial coordinates $(x_i, y_i)$ are extracted to form the foundation of the spatial representation.

Phenotypic identities (or clusters) are assigned through unsupervised k-means clustering ($K=10$) based on cellular morphology and marker expression. To build the cluster model, a random subset of 5 million cells is sampled from the total pool of $17{,}432{,}758$ cells. Each sampled cell is represented by a feature vector containing z-normalised marker intensities, cell area, and circularity. Following clustering, all cells in the dataset are assigned to their nearest cluster centroid, providing the discrete labels to be used as token number later.

To convert the irregular spatial distribution of cells into a structured format suitable for transformer models, we discretised each tissue core into a regular grid using a cell binning strategy. Let $\mathcal{P} = {(x_i, y_i)}_{i=1}^{N}$ denote the set of cell coordinates within a core. For each cell, we computed the Euclidean distance to its $k$-th nearest neighbour ($k=2$, excluding self-distance), and defined a characteristic local spacing as the median of these distances:
\begin{equation}
    d = \operatorname{median}_{i} \left( \lVert \mathbf{p}_i - \mathbf{p}_{i,(k)} \rVert \right).
\end{equation}

To ensure that bins capture local cellular neighbourhoods rather than individual pixels, a scaling factor $\alpha =2$ was applied, resulting the final bin size \(\Delta = \alpha d\). Each tissue core was partitioned into a uniform grid with a spacing or bin size of $\Delta$ (\(\Delta=0.8 \mu\text{m}\)), and each cell was assigned to its nearest grid bin. 

\begin{figure}[tb]
  \centering
  \includegraphics[width=\linewidth]{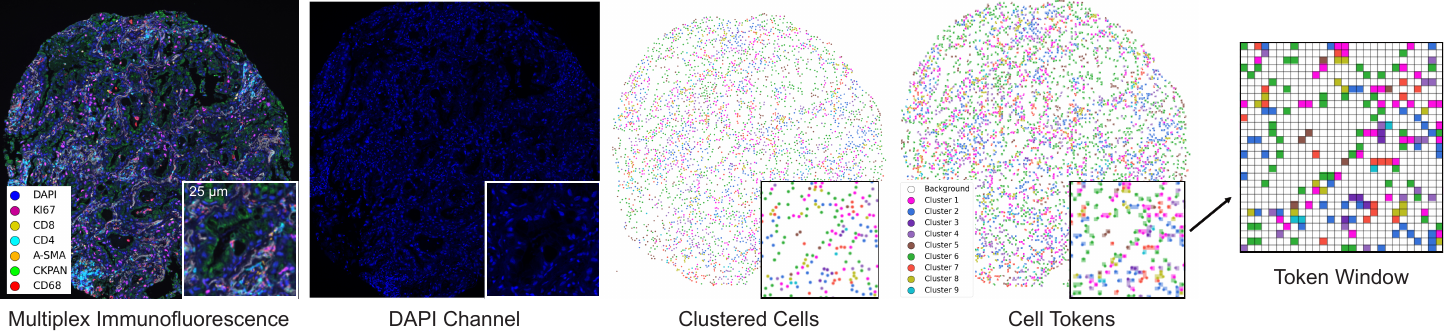}
  \caption{\textbf{Tokenisation of cellular spatial information from multiplex immunofluorescence images.} An example mIF image and its corresponding DAPI channel used for nuclear-based cell segmentation are shown. Extracted cell centroids are coloured according to their cluster assignments, representing discrete cellular phenotypes. These labelled cells are subsequently binned into a fixed-size spatial grid to generate cell tokens.}
  \label{fig:tokens}
\end{figure}

Each grid bin inherits the cluster number corresponding to the cell it contains and is treated as a discrete token. Grid bins without assigned cells are encoded using a dedicated background token. Analogous to whitespace in natural language, these tokens preserve spatial separation between neighbouring cell tokens and prevent the implicit collapse of spatial distances during sequence construction. This grid-based tokenisation preserves local biological structure while introducing minimal spatial approximation, enabling cell-level spatial information to be represented as sequences directly consumable by transformer architectures. A visual illustration of cell locations and their corresponding token assignments is shown in \Cref{fig:tokens}.

\subsubsection{Cross-dataset token assignment using shared markers}
\label{crossdatasetassignment}

To transfer similar cellular token identities from the reference dataset (primary) to the external dataset (additional), we employ a k-nearest neighbour (k-NN)-based label transfer strategy restricted to shared markers. Six biologically corresponding markers were used for alignment: CD4, CD8, CD68, Ki67, Cytokeratin, and a nuclear marker (DAPI in LATTICeA-IO and DRAQ5 in CTCL-P).

For computational efficiency and balanced token representation, a stratified subset of $25{,}000$ cells was sampled from the reference dataset, preserving the original phenotypic token proportions. For both datasets, mean marker intensities were extracted per cell and independently z-score normalised within the dataset. This independent normalisation avoids redefining cohort-specific expression contexts while ensuring comparable feature scaling across shared markers.

Token assignment was performed using a k-NN classifier ($k=250$), trained on the reference cells in the shared-marker feature space. For each target cell, the $250$ nearest reference neighbours were identified, and the cell was assigned the token receiving the majority of neighbour votes. Assignment confidence was defined as the fraction of neighbours voting for the assigned token:

\begin{equation}
  c_i = \frac{1}{k} \max_t \sum_{j=1}^{k} 1(y_j = t),
\end{equation}

where $y_j$ denotes the token identity of the $j$-th nearest neighbour.

The validation of cross-dataset tokenisation focuses on assignment confidence statistics and phenotypic consistency across cohorts. Specifically, we report the mean and median assignment confidence across all target cells, as well as token-wise confidence distributions.

\subsubsection{Masked Transformer Pretraining}

\begin{figure}[tb]
  \centering
  \includegraphics[width=\linewidth]{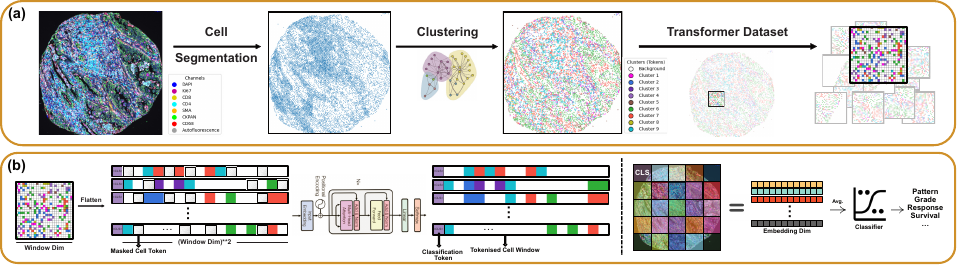}
  \caption{
  \textbf{MUL-T pipeline}
\textbf{(a)} Cells are segmented and then clustered using K-means. The tissue is then binned into a spatial grid where bins, acting as tokens, inherit majority cell cluster labels. These tokens are grouped within fixed spatial windows to create input sequences for the transformer. \textbf{(b)} A transformer model processes masked sequences to produce contextualised \texttt{[CLS]} representations for each sequence. These local spatial contexts are aggregated into core-level or patient-level vectors, enabling downstream predictions.}
  \label{fig:fig2}
\end{figure}

Following cell tokenisation, each tissue core is represented as a collection of integer-valued token sequences corresponding to inherited cell cluster numbers (\cref{fig:fig2}a). To preserve spatial structure within local contexts, tokens are augmented with rotary positional embeddings\cite{su2024roformer} within each window prior to transformer encoding. Each input sequence is constructed from a $30 \times 30$ grid of tokens, flattened into a one-dimensional sequence of length $900$. This window size was empirically selected to balance computational efficiency with the ability to capture meaningful spatial dependencies between neighbouring cells. For masked pretraining, we use a tiny BERT-style transformer encoder with 4 layers, 4 self-attention heads, and an embedding dimension of 256, resulting in 3.5M parameters. Owing to the small vocabulary size, the model remains lightweight and computationally efficient compared to standard transformer architectures. 

A random subset of token positions $\mathcal{M}$ is selected and replaced with a masking token, and the model is trained to recover the original identities from the surrounding context (\cref{fig:fig2}b). We explore masking ratios of 20-30\%. Background tokens are excluded from masking to concentrate learning on biologically meaningful structures. The resulting vocabulary consists of $13$ discrete tokens: $10$ phenotypic cluster tokens, one background token, one \texttt{[CLS]} token, and one masking token. The training objective is defined as the token-level cross-entropy loss restricted to masked positions:
\begin{equation}
    \mathcal{L}_{\mathrm{MLM}} = - \frac{1}{|\mathcal{M}|} \sum_{i \in \mathcal{M}} w_{y_i} \log p_\theta(y_i \mid \mathbf{x}),
\end{equation}

where $y_i$ denotes the ground-truth token at position $i$, $p_\theta(y_i \mid \mathbf{x})$ is the model-predicted probability, and $w_{y_i}$ denotes optional class-specific weighting (the proportion of training tokens belonging to each phenotypic cluster).

Optimisation was performed using AdamW with $\beta_1 = 0.9$, $\beta_2 = 0.95$, and a weight decay of $0.01$. The learning rate was set to $10^{-4}$ with a linear warm-up. Training was conducted with a batch size of $32$ per device using distributed data parallelism across two NVIDIA H200 GPUs (epoch time $\approx 12 $ minutes) for 300 epochs.

\subsubsection{Downstream Tasks}

After masked transformer pretraining, the model outputs a \texttt{[CLS]} embedding for each sequence (=window), summarising the local spatial context of the corresponding grid window. These \texttt{[CLS]} embeddings act as compact representations of local tissue architecture and form the basis for downstream prediction tasks at both tissue core and patient levels. Each \texttt{[CLS]} token therefore represents a spatial snapshot of tissue organisation within a local region of a patient sample.

For each tissue core, multiple spatial grid windows are extracted, resulting a set of \texttt{[CLS]} embeddings associated with that core. These embeddings correspond to spatially distinct windows sampled across the core, as illustrated in the second part of \cref{fig:fig2}b using a $5 \times 5$ set. A core-level representation is obtained by aggregating the \texttt{[CLS]} embeddings across all windows belonging to the same core. Similarly, for patients with multiple tissue cores, patient-level representations are constructed by aggregating embeddings across all their associated cores. Formally, let $\mathcal{W}c = {w_1, \dots, w_{N_c}}$ denote the set of spatial windows extracted from tissue core $c$, and let $\mathbf{z}^{\mathrm{CLS}}_{w} \in \mathbb{R}^{d}$ denote the \texttt{[CLS]} embedding produced by the transformer for window $w$, with $d$ (=transformer embedding dimension) . The core-level representation is defined as the mean aggregation of \texttt{[CLS]} embeddings:

\begin{equation}
    \mathbf{z}_{\text{core}}^{(c)} = \frac{1}{|\mathcal{W}_c|} \sum_{w \in \mathcal{W}_c} \mathbf{z}^{\mathrm{CLS}}_{w}.
\end{equation}
For patient $p$ with a set of associated tissue cores $\mathcal{C}p = {c_1, \dots, c_{M_p}}$, the patient-level representation is obtained by aggregating the corresponding core-level embeddings:

\begin{equation}
    \mathbf{z}_{\text{patient}}^{(p)} = \frac{1}{|\mathcal{C}_p|} \sum_{c \in \mathcal{C}_p} \mathbf{z}_{\text{core}}^{(c)}.
\end{equation}

This aggregation strategy results in fixed-dimensional representations suitable for downstream classifications using regular machine learning classifier models. Core-level prediction tasks include treatment response classification (responder vs.\ non-responder), PD-L1 positive, and predominant tumour pattern classification across two categories (Acinar, Cribriform, and Lepidic as lower risk patterns - Micropapillary, Papillary, and Solid as higher risk patterns). Patient-level predictions include tumour grade classification across three classes (G1, G2, and G3). For multi-class classification tasks, binary logistic regression was extended using a one-versus-rest strategy.

To demonstrate performance, we compare against two common baseline approaches, as well as the foundational ViT model, KRONOS \cite{shaban2024foundation}. For KRONOS, we extract features from 256-pixel tiles within each core and aggregate these embeddings in the same way as our \texttt{[CLS]} representations, ensuring a controlled comparison with MUL-T. In addition, we evaluate two classical feature-based baselines. First, marker intensity statistics, where per-marker mean and standard deviation are computed across cells within each core. Second, cluster frequency features, where the normalised proportion of cells assigned to each K-means phenotypic cluster is used.

\section{Experiments}

This section evaluates the MUL-T framework from two complementary perspectives. First, we assess the quality of the tokenisation stage by analysing clustering of the primary dataset and token assignment of the additional dataset. Second, we evaluate the performance of the proposed method for downstream prediction and benchmark it against two baselines and a state-of-the-art foundation model. These experiments target both core-level and patient-level tasks, including treatment response, PD-L1 positivity, tumour pattern, and grading. 

\subsection{Tokenisation} 

\begin{figure}[tb]
  \centering
  \includegraphics[width=\linewidth]{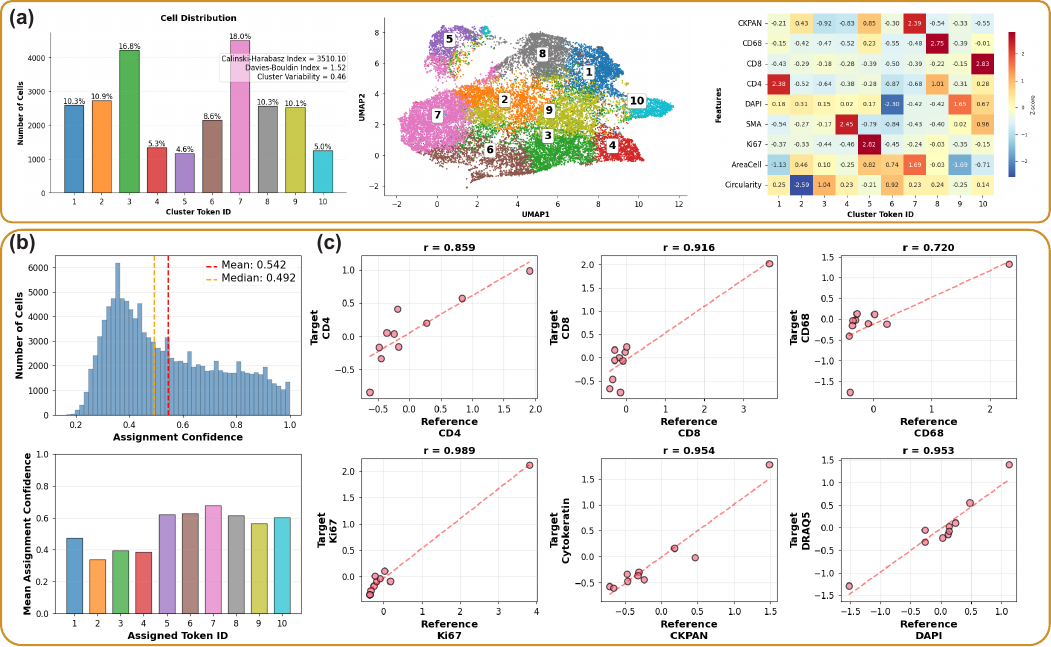}
  \caption{\textbf{Assessment of clustering, tokenisation, and cross-dataset marker alignment.} \textbf{(a)} Distribution of cluster tokens and corresponding feature profiles in the primary dataset (LATTICeA-IO). \textbf{(b)} Assignment confidence for cross-dataset token transfer. \textbf{(c)} Correlation of shared markers between the primary and additional cohorts.}
  \label{fig:clustering}
\end{figure}

Using K-means clustering in the primary dataset, we identified 10 distinct clusters (token IDs). The cell features (normalised marker intensities and cell shapes) are not fully independent, leading to partially overlapping cluster profiles. Nevertheless, all clusters display statistically significant differences across features (\(p < 0.05\)). The clusters are both compact and well separated (BCSS/WCSS=1.27). They also maintain balanced cell distributions (see bar plot in \cref{fig:clustering}a.) A complementary UMAP visualisation of a subsample of 25,000 cells, colour-coded by cluster, further illustrates this separation.
The heatmap of cluster-level feature profiles presents correlations between cell features and cluster numbers. Some clusters show strong positive or negative associations with specific markers and morphological traits. For example, cluster 1 groups cells with high CD4 expression but smaller areas and lower SMA and CD8 levels. In contrast, cluster 2 captures more elongated, spindle-shaped cells with higher cytokeratin expression.

A k-nearest-neighbour-based (\(k=250\)) token transfer approach resulted in a robust assignment of the reference dataset clusters to the additional dataset cells. A confidence score for each cell is
defined as the fraction of the nearest neighbours' votes for the assigned token. Across all target cells, the mean assignment confidence score is reported as 0.54, ranging from 0.12 to 1.00. (\Cref{fig:clustering}b, per cell histogram plot). This high-confidence assignment is observed for \(49.2\%\) of the target cells. Assignment confidence varied across tokens (\Cref{fig:clustering}b, per token bar plot), with the mean per token confidence ranging from 0.338 to 0.676.

Despite substantial biological differences in cellular composition between the primary and additional datasets, transferred tokens showed strong biological consistency at the marker level. Token-wise mean marker intensities were highly correlated between the reference and target datasets across all six shared markers (\Cref{fig:clustering}c), with a mean Pearson correlation coefficient of 0.8545 and values ranging from 0.720 to 0.989. These results demonstrate that label transfer preserves token-specific molecular profiles across datasets, supporting the biological validity of the transferred token identities under domain shift.

\subsection{Downstream Tasks}

\begin{figure}[tb]
  \centering
  \includegraphics[width=1\textwidth]{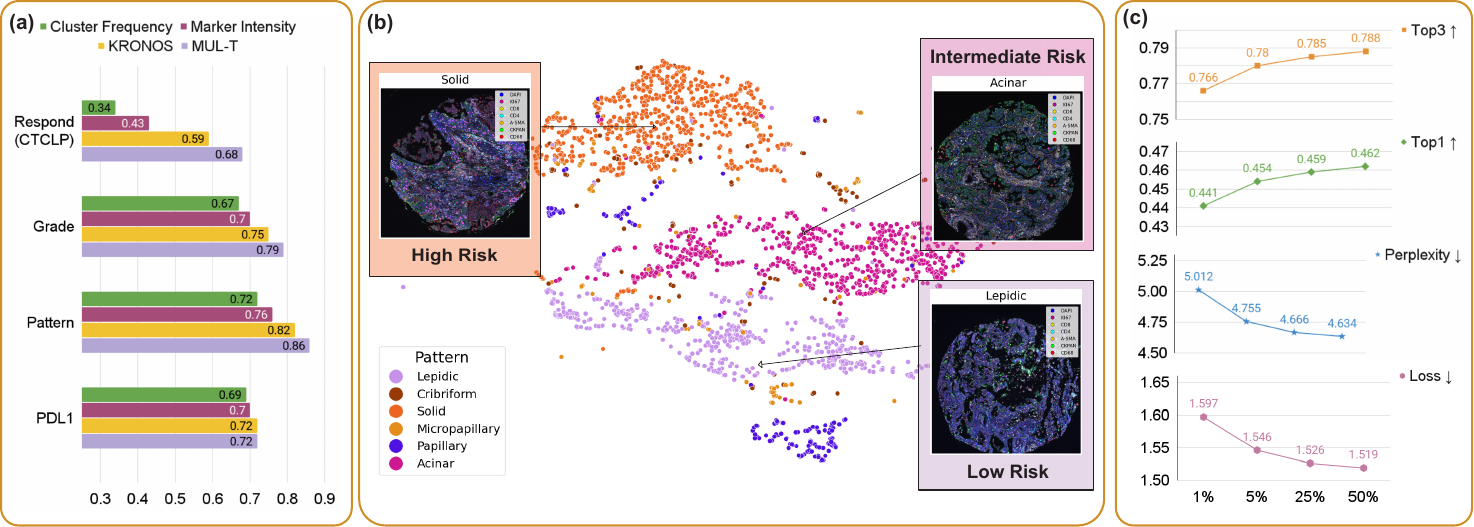}
  \caption{\textbf{Experimental evaluation.} \textbf{(a)} Benchmarking of downstream tasks against classical baselines and a foundation model. \textbf{(b)} UMAP projection of core-level embeddings coloured by dominant tumour pattern. \textbf{(c)} Data scalability analysis, reporting loss, perplexity, and Top-1 and Top-3 accuracy for masked token prediction during transformer pretraining.}
  \label{fig:abliation}
\end{figure}

Using the generated \texttt{[CLS]} embeddings, we evaluate clinically relevant downstream tasks at both core and patient levels. In the primary dataset, we assess PD-L1 positivity and predominant pattern classification at the core level, and grading at the patient level. Cross-dataset generalisation is further examined using zero-shot \texttt{[CLS]} embeddings from the additional CTCL-P cohort for treatment response prediction (\Cref{fig:abliation}a).

The model achieves an AUC of 0.86 for pattern classification, 0.79 for patient grading, and 0.72 for PD-L1 positivity. Without task-specific training, it differentiates responders from non-responders in the external cohort with an AUC of 0.68. Across all tasks, the model outperforms classical baselines and achieves performance comparable to the foundation model. All results are validated using 10-fold cross-validation, and AUC scores are reported using balanced averaging to account for class imbalance.

To further support core-level tumour pattern prediction, we perform a retrieval-based evaluation. The model attains a ranking accuracy of 0.56 across six imbalanced pattern classes and a mean average precision of 0.29 when measuring label consistency among the top-$k$ neighbours ($k=250$). These results highlight the suitability of transformer-based \texttt{[CLS]} embeddings for clinically meaningful similarity search between cores. \Cref{fig:abliation}b shows a supervised UMAP projection of averaged core-level embeddings coloured by dominant histological pattern, illustrating clear separation induced by the learned representations.

Furthermore, we examine data scalability and observe consistent but gradually diminishing improvements as the training set increases from 1\% to 50\%. Loss decreases from 1.597 to 1.519 and perplexity from 5.01 to 4.63, while Top-1 accuracy increases modestly from 0.441 to 0.462 (\Cref{fig:abliation}c). Most gains occur at very small data fractions, after which performance stabilises. This suggests that the model rapidly captures the dominant statistical structure of the cellular token space.

The early saturation likely reflects limited model capacity rather than insufficient data. Despite its small size, the transformer achieves competitive accuracy with substantially fewer parameters and lower training cost. Overall, these results indicate that the cellular token space contains structured regularities that can be effectively learned by a lightweight architecture, offering a favourable balance between efficiency and representational performance.

\section{Discussion}
The MUL-T framework represents a shift in how we interpret multiplex tissue images, moving away from global phenotype aggregation towards a context-aware approach. By treating cells and their spatial arrangements as fundamental units, the model captures the grammar of tissue architecture.

A key strength of this framework is the use of masked cell phenotype modelling through the attention mechanism. By tasking the model with inferring missing cell tokens from their neighbours, we force it to learn the underlying biological rules of the microenvironment. This mirrors the biological reality where cellular function is not an isolated trait but is continuously modulated by surrounding signals. Unlike traditional methods that rely on simple marker intensities, MUL-T identifies higher-order spatial motifs that may be indicative of specific disease states or immune responses.

The technical implementation of grid-based tokenisation further bridges the gap between spatial fidelity and computational practicality. By deriving grid resolution from empirical inter-cell distances, the framework avoids the computational overhead of graph neural networks while maintaining the integrity of the tissue layout. Our findings suggest that phenotype homogeneity is preserved within these bins, ensuring that biological signals are not lost during discretisation. Furthermore, the inclusion of background tokens is a critical design choice.This acknowledges that empty space in a tissue sample is structurally informative and prevents the model from creating artificial adjacencies between distant cell clusters.

A core strength of the proposed framework is its streamlined architecture, which utilises a tiny BERT-style transformer encoder. Because the vocabulary of cell phenotypes is significantly smaller than that of human natural language, the model requires substantially fewer parameters. This makes the architecture exceptionally lightweight and computationally efficient compared to standard transformer models. This design prevents over-parameterisation while maintaining the capacity to learn complex relational and spatial dependencies within the tissue microenvironment.

Finally, the decision to use a lightweight machine learning downstream model, such as logistic regression, and average aggregation serves as a rigorous validation of the learned representations. It ensures that the predictive power is derived from the quality of the embeddings themselves rather than the complexity of the classifier. This approach enhances interpretability which is a vital requirement for clinical adoption.

\subsection{Limitations}

Despite MUL-T's strengths, the proposed framework has several limitations. First, while the grid-based tokenisation preserves local spatial structure, it introduces a discretisation step that may smooth fine-grained spatial variations, particularly in regions with highly heterogeneous cell packing or rare phenotypes. Although empirical analysis suggests minimal phenotype mixing within bins, extreme cellular densities or atypical tissue architectures may violate this assumption. In addition, the model focuses on local neighbourhoods and does not explicitly model long-range spatial dependencies, which may be relevant for certain global tissue-level processes.

Second, the current formulation relies on predefined cell phenotypes derived from upstream clustering or annotation, making performance partially dependent on the quality and consistency of cell segmentation and phenotype assignment (clustering). Errors at this stage may propagate into downstream representations. While the model supports cross-dataset transfer through marker alignment, complete biological equivalence between cohorts cannot be guaranteed, particularly when marker panels or staining protocols differ substantially. Finally, although lightweight downstream models improve interpretability and stability, they may limit performance ceilings compared to task-specific end-to-end optimisation.

\section{Conclusion}
MUL-T introduces a scalable and biologically grounded approach to multiplexed image analysis. By reframing tissue architecture as a contextual prediction problem, the framework successfully extracts complex spatial relationships that are often missed by traditional global metrics.

The results demonstrate that self-supervised masked modelling can generate transferable embeddings that align with histopathological logic. Despite the challenges posed by upstream data variability and the local nature of the attention windows, the framework provides a transparent and interpretable pathway for clinical data analysis. Future work may explore the integration of multi-scale attention mechanisms to bridge the gap between local microenvironments and global tissue morphology. Ultimately, MUL-T establishes a foundation for applying the principles of large language models to the intricate language of tissue.


\section*{Acknowledgements}
The authors would like to express their sincere gratitude to the group's pathologists for their guidance on cell biology and marker panel design, which were instrumental in ensuring the clinical relevance and biological integrity of this work.

%
%
\bibliographystyle{splncs04}
\bibliography{main}
\end{document}